# Text classification using machine learning methods

**Bogdan Oancea**[1*]

[1]University of Bucharest, Faculty of Business and Administration, Department of Applied Economics and Quantitative Analysis, Bucharest, Romania

**Abstract:** In this paper we present the results of an experiment aimed to use machine learning methods to obtain models that can be used for the automatic classification of products. In order to apply automatic classification methods, we transformed the product names from a text representation to numeric vectors, a process called word embedding. We used several embedding methods: Count Vectorization, TF-IDF, Word2Vec, FASTTEXT, and GloVe. Having the product names in a form of numeric vectors, we proceeded with a set of machine learning methods for automatic classification: Logistic Regression, Multinomial Naive Bayes, kNN, Artificial Neural Networks, Support Vector Machines, and Decision trees with several variants. The results show an impressive accuracy of the classification process for Support Vector Machines, Logistic Regression, and Random Forests. Regarding the word embedding methods, the best results were obtained with the FASTTEXT technique.

## 1. Introduction

The modernization strategy of official statistics includes the adoption of new methods used in statistical processes and the integration of new data sources in statistical production. Thus a few years ago we started the process of collecting online prices from the main national e-commerce sites, using webscraping techniques in order to use them for Consumer Price Index computation (Oancea and Necula, 2019). The volume of data collected through webscraping is very large: approximately 50,000 records collected each week. In order to be used in a statistical production process, products must be classified according to the categories used for CPI calculation. Manual labelling of these records is almost impossible due to the large volume of data and for this reason we experimented with an automatic classification process, starting with a small set of product categories. Thus, we have chosen 15 categories from the ECOICOP international classification, from the food and household appliances categories.

## 2. Methods

Experimentation with the automatic classification process began with the random selection of a sample of approx. 2500 products, and we proceeded to manually label these products according to the international ECOICOP classification. Next, in order to be able to use different machine learning techniques for automatic classification, we transformed the product names, which are text-type data, into numerical vectors, a technique called word embedding coming from the field of Natural Language Processing. We used different techniques to build vectorization of product names: Count Vectorization (Sparck, 1972), TF-IDF (Ullman and Leskovec, 2014), Word2Vec (Mikolov et al., 2013), FASTTEXT (Joulin et al., 2016) and GloVe (Pennington et al., 2014).

---

[*] Corresponding author's email: "Place corresponding author's email here"

**Count Vectorization** is a simple technique that builds a vocabulary of words that appear in product names, and then builds the number vector corresponding to each name by the frequency of occurrence of each word in the vocabulary.

**Term frequency – Inverse document frequency (TF-IDF)** basically consists in the calculation of two indicators:

- The normalized frequency of a term (word): measures how often a term/word appears and is calculated as the ratio between the number of occurrences of a word in a document and the total number of words in the document (in our case document = product name ):

$$Tf(t,d) = \frac{n_{t,d}}{\sum_{t'} n_{t'd}} \quad (1)$$

where *t* is the term (word) and *d* is the document where this term appears

- Inverse of document frequency: measures how important a term is and is calculated as the logarithm of the ratio between the total number of documents and the number of documents in which the term of interest appears:

$$idf(t,V) = \log(\frac{N}{|\{d \in V; t \in d\}|}) \quad (2)$$

where *t* is the term (word), *V* - the vocabulary and *N* the total number of documents.

The calculation formula for TF-IDF will be:

$$tfidf(t,d,V) = tf(t,d) \times idf(t,V) \quad (3)$$

Both in the case of using Count Vectorization and in the case of TFIDF, when building the vocabulary we took into account, in addition to the actual words, also constructions of the n-gram type, i.e. sequences of n words that appear in the text. Specifically, we built all n-gram combinations with up to 3 consecutive words.

**Word2Vec** is an algorithm that takes as input a vocabulary (a set of words) and outputs a vector (numerical) representation for each word in the vocabulary).

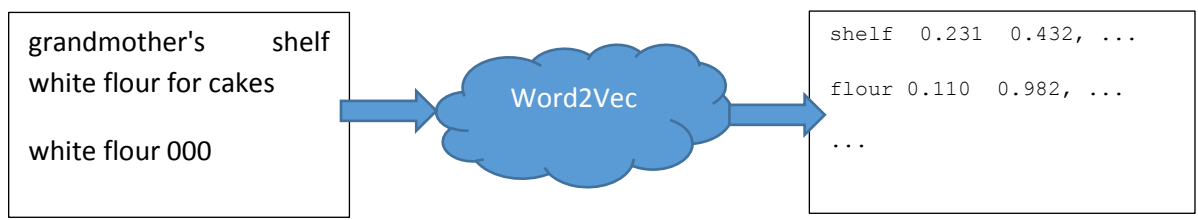

**Figure 1: Word2Vec vectorization**

There are two versions of this algorithm: CBOW (Continuous Bag of Words) which starts from a word in the text, considers the words around it (the context) and tries to predict that word and SKIP-GRAM which starts from a word and tries to predicts the words that appear in the text around it (context words). Both versions use an artificial neural network and the vector representation of the words will be given by the weights of the resulting network after the training process. The activation function of the hidden layer of the network is linear and the output function is softmax.

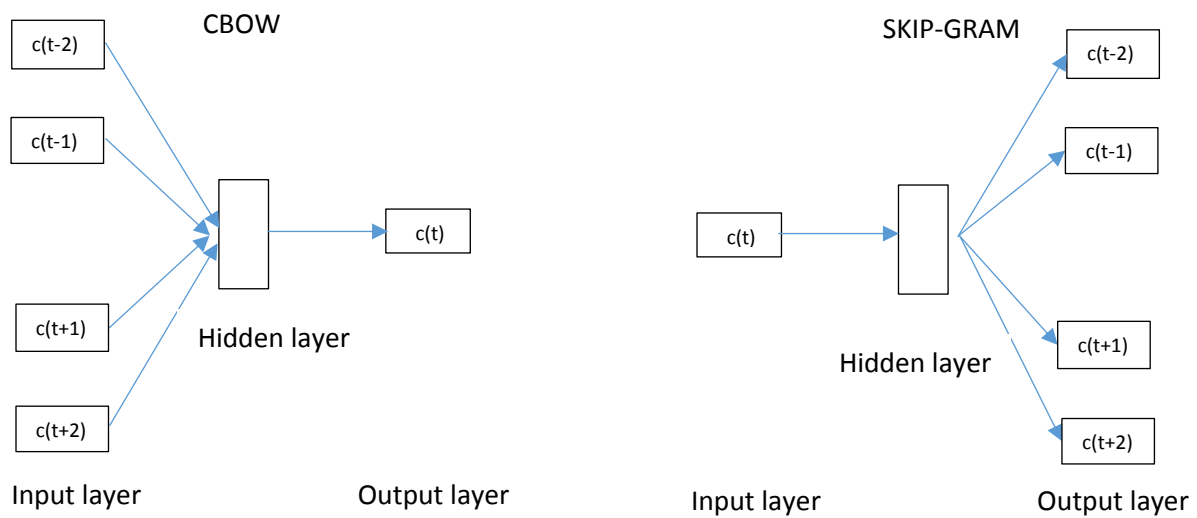

Figure 2: CBOW and SKIP-GRAMK versions of Word2Vec

On the right side, the CBOW version with a window of 4 words is represented schematically: the input of the neural network will be represented by the 4 context words and the output by the target word. After the recipe training process, the vector representation of each target word will be given by the weights of the artificial neural network.

On the left we have the SKIP-GRAM version of the Word2Vec algorithm, where the neural network is presented with a word as input and outputs the context of that word (through a window of 4 words).

**FASTTEXT** is an algorithm developed by the Facebook AI Research Laboratory and has the same two versions as Word2Vec: CBOW and SKIP-GRAM, also using an artificial neural network, but in the training process it uses both words from the dictionary and n-grams (pairs of n letters that make up words). In this way, FASTTEXT can also handle the case of words that do not appear in the initial network training dictionary.

**GloVe** is an unsupervised learning algorithm for building vector representations of words. The model is trained on global aggregated word-to-word co-occurrence statistics from a vocabulary. The global word-word co-occurrence matrix, counts the frequency with which words appear with each other in a given vocabulary (set of words). Populating this array requires only one pass through the entire vocabulary and is done only once. GloVe is essentially a log-bilinear model with a weighted least squares objective function.

For Word2Vec we built the vectorization of product names in two different ways:

- by adding the vector representations of each word that make up the product name.
- by averaging the vectorizations of each word.

For FASTTEXT, the vectorization of product names was obtained by dividing the vectors of the component words by their L2 norm and calculating the average value only for the vectors with L2 different from zero.

In the case of GloVe we just used the addition of the vector representations of each word that make up the product name to get the vector representation of the name.

For building product name vectorization, we limited ourselves to 3000 dimensions for the Count Vectorization and TF-IDF methods and 50 for Word2Vec, FASTTEXT, and GloVe to keep the runtime within acceptable limits for this experiment. For Count Vectorization and TF-IDF, we considered not only words that make up product names, but also n-grams with a maximum of 3 words. For the Word2Vec and FASTTEXT methods, we applied both the CBOW and SKIP-GRAM approaches. Thus, we've built 9 different vectorizations for each product.

Having the numerical representation of the product names, we continued with a series of supervised learning techniques for automatic classification:

- Multinomial logistic regression (Böhning, 1992)
- Naive Bayes classifier (Xu, 2018)
- Decision trees in the Gini Index and Informational Gain versions to decompose tree nodes (Wu et al., 2010)
- Bagged trees (Abellan and Masegosa, 2010)
- Decision trees C4.5 (Quinlan, 2014)
- C50 decision trees (Kuhn and Quinlan, 2023)
- Support Vector Machines (SVM) with radial and sigmoid kernel (Cortes and Vapnik, 1995)
- Random forests (Ho, 1995)
- K-Nearest Neighbors (Mucherino et al., 2009)
- Artificial Neural Networks (ANN) (Haykin, 2009)
- eXtreme Gradient Boosted Trees (XGBoost) (Chen and Guestrin, 2016).

After dividing the data set into training and testing subsets, we trained each model mentioned above on the training data set and then applied the model to the test data sets. We even implemented a grid search procedure to choose the optimal values for the classifier parameters. However, the grid search procedure is time-consuming and requires the use of special parallel programming techniques to keep the running time reasonable. For this experiment, we used the grid search procedure only for SVM, XGBoost, KNN and ANN. We also used a cross validation technique (10-fold cross validation).

## 3. Results

For each classification method, we calculated:

- Confusion matrix (confusion matrix)
- Accuracy (number of correctly classified products / total number of products)
- F1 score (used especially if there is an imbalance between classes)

The F1 score was calculated in two ways: as a simple average and as a weighted average of the F1 score for each class. The weights were calculated as (1-class frequency), to mitigate the pronounced imbalance of the number of products in the considered classes.

The results showed an impressive accuracy of the automatic classification with values between 0.76 and 0.99. The best classifiers were found to be Random Forests, ANN and SVM with radial kernel. In terms of vectorization methods, the best results were obtained with Count Vectorization, TF-IDF, GLOVE, and FASTTEXT, while Word2Vec showed the lowest accuracy for almost all classification methods used in our study. All data processing was performed using the software system R ver. 4.2. on an Intel Core i7-8559U processor at 4.5 GHz, 32 GB DDR4 RAM, and a Windows 11 operating system

Table 1 shows all ML methods tested together with the word vectorization technique that gave the highest accuracy. A more detailed description of the results and a discussion on the performance of each method can be found in (Oancea, 2023).

**Table 1: Performance metrics for automatic product classification**

| Classification method | Vectorization method | Accuracy | F1 | Weighted F1 |
|---|---|---|---|---|
| Logistic regression | Count vectorization | 0.976 | 0.924 | 0.859 |
| Naïve Bayes | Count vectorization | 0.989 | 0.943 | 0.877 |
| Decision trees (Gini) | TF-IDF | 0.963 | 0.961 | 0.882 |
| Decision tress (information) | TF-IDF | 0.952 | 0.937 | 0.852 |
| Bagged decision trees | TF-IDF | 0.983 | 0.945 | 0.879 |
| C4.5 | Count vectorization | 0.984 | 0.945 | 0.880 |
| C50 | Count Vectorization | 0.984 | 0.946 | 0.881 |
| Random forests | GLOVE | 0.991 | 0.971 | 0.904 |
| KNN | TF-IDF | 0.987 | 0.945 | 0.880 |
| ANN | GLOVE | 0.991 | 0.975 | 0.906 |
| SVM (radial kernel) | FASTTEXT SKIP GRAM | 0.997 | 0.988 | 0.922 |
| SVM (sigmoid kernel) | FASTTEXT SKIP GRAM | 0.982 | 0.931 | 0.861 |
| XGBoost | Count Vectorization | 0.986 | 0.945 | 0.880 |

4. **Conclusions**

The results obtained are very encouraging, the automatic classification methods tested so far showing very good performance. These performances can be explained also by the fact that product names do not vary greatly from one retailer to another. Once a classification model is applied to a training data set, the test data provided for classification largely follow the same rules for building product names, and the classification will show good results.

However, there are still some limitations that could be addressed in the future. The automatic classification algorithms were run on a small data set, and yet the computation time was high, especially for the methods where grid search techniques were used in order to choose the optimal parameter values. This limitation will be accentuated when we try to classify larger datasets and special parallel programming techniques should be considered.

Another issue we should consider for the future is out-of-vocabulary (OOV) words. While the FASTTEXT word vectorization method can handle such words, the other methods cannot handle OOV words. A strategy should also be devised to mitigate this limitation.

The automatic classification technique has various applications, not only in the field of product names. For example, a statistical survey where the occupation code of the respondent is needed, may ask him/her to enter a description of his occupation in free terms and then using classification techniques to choose the code according to the nomenclature in force.